\pdfoutput = 1
\documentclass{article}

\usepackage[utf8]{inputenc} % allow utf-8 input
\usepackage[T1]{fontenc}    % use 8-bit T1 fonts
\usepackage{lmodern}
\usepackage{natbib}
\usepackage{hyperref}       % hyperlinks
\usepackage{url}            % simple URL typesetting
\usepackage{booktabs}       % professional-quality tables
\usepackage{amsfonts}       % blackboard math symbols
\usepackage{nicefrac}       % compact symbols for 1/2, etc.
\usepackage{microtype}      % microtypography
\usepackage{booktabs,caption}
\usepackage[flushleft]{threeparttable}
\usepackage{authblk}
\usepackage{fullpage}
\usepackage{multicol}
\usepackage{pifont}% http://ctan.org/pkg/pifont
\usepackage{wrapfig}
\newcommand{\cmark}{\ding{51}}%
\newcommand{\xmark}{\ding{55}}%

\usepackage{graphicx,color}
\usepackage{epsfig}
\usepackage{amsmath,amsthm,amssymb}
\usepackage{subfig}
\usepackage{multirow}
\usepackage[linesnumbered,ruled]{algorithm2e}
\newtheorem{thm}{Theorem}
\newtheorem{assumption}{Assumption}

\newtheorem{lem}{Lemma}
\newtheorem{prop}{Proposition}
\newtheorem{defn}{Definition}

\theoremstyle{remark}
\newtheorem{rem}{Remark}

\graphicspath{{figures/}}
\usepackage{mdwlist}

\title{Online Robust Policy Learning in the Presence of Unknown Adversaries}

\author{
   Aaron J. Havens,\;Zhanhong Jiang,\;Soumik Sarkar\\
  Department of Mechanical Engineering\\
  Iowa State University\\
  Ames, IA 50011 \\
  \texttt{\{ajhavens,zhjiang,soumiks\}@iastate.edu} \\
}

\begin{document}

\maketitle

%%%%%%%%%%%%%%%%%%%%%%%%%%%%%%%%%%%%%%%%%%%%%%%%%%%%%%%%%%%%%%%%%%%%%%%%%%%%%%%%
\begin{abstract}
The growing prospect of deep reinforcement learning (DRL) being used in cyber-physical systems has raised concerns around safety and robustness of autonomous agents. Recent work on generating adversarial attacks have shown that it is computationally feasible for a bad actor to fool a DRL policy into behaving sub optimally. Although certain adversarial attacks with specific attack models have been addressed, most studies are only interested in off-line optimization in the data space (e.g., example fitting, distillation). This paper introduces a Meta-Learned Advantage Hierarchy (MLAH) framework that is attack model-agnostic and more suited to reinforcement learning, via handling the attacks in the decision space (as opposed to data space) and directly mitigating learned bias introduced by the adversary. In MLAH, we learn separate sub-policies (nominal and adversarial) in an online manner, as guided by a supervisory master agent that detects the presence of the adversary by leveraging the \textit{advantage} function for the sub-policies. We demonstrate that the proposed algorithm enables policy learning with significantly lower bias as compared to the state-of-the-art policy learning approaches even in the presence of heavy state information attacks. We present algorithm analysis and simulation results using popular OpenAI Gym environments.

\end{abstract}

%%%%%%%%%%%%%%%%%%%%%%%%%%%%%%%%%%%%%%%%%%%%%%%%%%%%%%%%%%%%%%%%%%%%%%%%%%%%%%%%
\section{Introduction}
Real applications of cyber-physical systems that utilize learning techniques are already abundant such as smart buildings~\cite{shih2016designing}, intelligent transportation networks~\cite{rawat2015towards}, and intelligent surveillance and reconnaissance~\cite{antoniou2016general}. In such systems, Reinforcement Learning (RL)~\cite{sutton1992reinforcement,sutton2017reinforcement} is becoming a more attractive formulation for control of complex and highly non-linear systems. The application of Deep Learning (DL) has pushed recent advances in RL, namely Deep RL (DRL)~\cite{mnih2015human,mnih2016asynchronous,van2016deep}. Particularly in 3D continuous control tasks, DL is an indispensable tool due to its ability to generalize high dimensional state-action spaces in Policy Optimization algorithms \cite{lillicrap2015continuous}, \cite{levine2016end}. Notable variance reduction and trust-region optimization strategies have only furthered the performance and stability of DRL controllers~\cite{schulman2015trust}. 

Although DL is generally useful for these control problems, DL has inherent vulnerabilities in the way that even very small perturbations in state inputs can result in significant loss in policy learning performance. This becomes a very reasonable cause for concern when contemplating DRL controllers in real-world tasks where there exist, not only environmental uncertainty, but perhaps adversarial actors that aims to fool a DRL agent into making a sub-optimal decision. During policy learning, information perturbation can be generally thought of as a bias that can prevent the the agent from effectively learning the desired policy. Previous attempts in mitigating adversarial attacks have been successful against specific attack models, however, such robust training strategies are typically off-line (e.g., using augmented datasets~\cite{madry2017towards}) and may fail to adapt to different attacker strategies in an online fashion. Recently ~\cite{lin2017detecting} has taken a model-agnostic approach by predicting future states, however it may be susceptible to multiple consecutive attacks.

\textbf{Contributions}: In this paper, we consider a policy learning problem where there are periods of adversarial attacks (via corrupting state inputs) when the agent is continuously learning in its environment. Our main objective is online mitigation of the bias introduced into the nominal policy by the attack. We only consider how an attack affects the return instead of optimizing the observation space. In this context, our specific contributions are:
\begin{enumerate*}
\item \textbf{Algorithm} We propose a new hierarchal meta-learning framework, MLAH that can effectively detect and mitigate the impacts of adversarial state information attacks in a attack-model agnostic manner, using only the advantage observation.
\item \textbf{Analysis}: Based on a temporal expectation definition, we analyze the performance of a single mapping policy and our proposed multi-policy mapping. Visitation frequency estimates leads us to obtaining a new pessimistic lower bound for TRPO and variants.
\item \textbf{Implementation}: We implement the framework in widely utilized Gym benchmarks~\cite{brockman2016openai}. It is shown that MLAH is able to learn minimally biased polices under frequent attacks by learning to identify the adversaries presence in the return.
\end{enumerate*}
Although we mention several relevant techniques on learning with adversaries, we only contrast methodologies in table \ref{table_compare} that aim to mitigate adversarial attacks, as other papers ~\cite{pattanaik2018robust},~\cite{pinto2017robust} do not claim to do so. We compare our results with the state-of-the-art PPO~\cite{schulman2017proximal} that is sufficiently robust to uncertainties to understand the gain from multi-policy mapping.
\setlength\tabcolsep{2pt}
\begin{table}
\begin{center}
\begin{threeparttable}
\caption{Comparisons with different robust adversarial RL methods}\label{table_compare}
  \begin{tabular}{c c c c c c}
    \hline
    Method & Online & Adaptive& Attack-model agnostic & Mitigation\\ \hline
    VFAS~\cite{lin2017detecting} & \cmark & \xmark & \cmark & \cmark\\
    ARDL~\cite{madry2017towards} & \xmark & \xmark & \xmark  & \cmark\\ 
    {MLAH [This paper]} & \cmark & \cmark & \cmark & \cmark\\
    \hline
  \end{tabular}

\begin{tablenotes}
    \small
    \item Online: no offline training/retraining required, Adaptive: can adapt to a change in attack strategy, Attack-model agnostic: assumes no specific attack model, Mitigation : is the impact of the attack actively mitigated?
  \end{tablenotes}\vspace{-10pt}
\end{threeparttable}
\end{center}
\end{table}

\textbf{Related work}: Attacks on deep neural networks and mitigation strategies have only recently been studied primarily for supervised classification problems. These attacks are most commonly formulated as first order gradient-based attacks, first seen as FGSM by Goodfellow et al~\cite{goodfellow6572explaining}. These gradient based perturbation attacks have proven to be effective in misclassification, with the corrupted input often being indistinguishable from the original. The same principle applies to DRL agents, which can drastically affect the agent performance and bias the policy learning process. The authors in~\cite{huang2017adversarial} showed a threat model that considered adversaries capable of dramatically degrading performance even with small adversarial perturbations without human perception. Three new attacks for different distance metrics were introduced in~\cite{carlini2017towards} in finding adversarial examples on defensively distilled networks. The authors in~\cite{kos2017delving} introduced three new dimensions about adversarial attacks and used the policy's value function as a guide for when to inject perturbations. Interestingly, it has been seen that training DRL agents on designed adversarial perturbations can improve robustness against general model uncertainties~\cite{pinto2017robust}, \cite{pattanaik2018robust}. The adversarial robust policy learning algorithm~\cite{mandlekar2017adversarially} was introduced to leverage active computation of physically-plausible adversarial examples in the training period to enable robust performance with either random or adversarial input perturbations. Another robust DRL algorithm, EPOpt-$\epsilon$ for robust policy search algorithm~\cite{rajeswaran2016epopt} was proposed to find a robust policy using the source distribution. Note that the recently mentioned methods do not aim to mitigate adversarial attacks at all, but intentionally bias the agent to perform better for model uncertainties. 
%For more details of the comparison with the state-of-the-art, please see the table~\ref{table_compare}. 
%\subsection{Background and main contributions}
%%%%%%%%%%%%%%%%%%%%%%%%%%%%%%%%%%%%%%%%%%%%%%%%%%%%%%%%%%%%%%%%%%%%%%%%%%%%%%%%
\section{Preliminaries and Problem Formulation}
In this paper, we consider a finite-horizon discounted Markov decision processes (MDP), where each MDP $m_i$ is defined by a tuple $M=(\mathcal{S}, \mathcal{A}, \mathcal{P}, r, \gamma, \rho_0)$ where $\mathcal{S}$ is a finite set of states, $\mathcal{A}$ is a finite set of actions, $\mathcal{P}$ is a mapping function that signifies the transition probability distribution, i.e., $\mathcal{S}\times\mathcal{A}\times\mathcal{S}\to\mathbb{R}$, $r$ is a reward function $\mathcal{S}\to\mathbb{R}$ with respect to a given state and $r\in[r_{min}, r_{max}]$, $\rho_0$ is a distribution of the initial states and $\gamma\in(0,1)$ is the discounted factor.
The finite-horizon expected discounted reward $\mathcal{R}(\pi)$ following a \textit{policy} $\pi$ is defined as follows:
\begin{equation}\label{eq1}
\mathcal{R}(\pi)=\mathbb{E}_{s_0,a_0,...}\bigg[\sum_{t=0}^{T}\gamma^tr(s_t)\bigg]
\end{equation}
where $s_0\sim\rho_0(s_0), a_t\sim\pi_i(a_t|s_t), s_{t+1}\sim\mathcal{P}(s_{t+1}|s_t,a_t)$. We want to maximize this discounted reward sum by optimizing a policy $\pi : \mathcal{S} \rightarrow \mathcal{A}$ map, discussed next.
\subsection{Trust Region Optimization for Parameterized Policies}
For more complex 3D control problems, policy optimization has been proven to be the state-of-the-art approach. A multi-step policy optimization scheme presented in~\cite{schulman2015trust} dually maximizes the improvement (Advantage function) of the new policy while penalizing the change between the old and new policy described by a statistical distance, namely the Kullback Liebler divergence.
For continuous control policy optimization a variant of the advantage function is often used being the \textit{Generalized Advantage Function} (GAE) from ~\cite{schulman2015high}, which is parameterized by $\gamma$ and $\lambda$ where $V(s_t)$ is the value function. Intuitively, GAE attempts to balance the trade-off between bias and variance in the advantage estimate by introducing the controlled parameter $\lambda$. We will use this in policy optimization as well as a method for temporal state abstraction later in the proposed algorithm.
\begin{equation}\label{adv}
A_{GAE,t} = \zeta_t + (\gamma\lambda)\zeta_{t+1}+...+
...(\gamma\lambda)^{T-t+1}\zeta_{T-1}
\end{equation}
where $\zeta_t = r_t+\gamma V(s_{t+1})-V(s_t)$, $\gamma, \lambda\in[0,1]$.
% The KL divergence can be practically emulated by clipping the policy probability ratio about $1$ by some $\sigma \in [0,1]$. This clipped surrogate loss, $L^{CLIP}$, selects a minimum over the clipped ratio advantage and nominal ratio advantage to serve as a pessimistic lower-bound as suggested by Schulman et al~\cite{schulman2017proximal}. Let $r_t(\theta)$ be the policy ratio between the old and new policy network parameters.
% \begin{align*}
% r_t(\theta) = \frac{\pi(a_t | s_t)_{\theta}}{\pi(a_t | s_t)_{\theta_{old}}}
% \end{align*}
% Therefore, we have
% \begin{equation}
% \begin{aligned}
% L^{CLIP}(\theta) &= \E[\text{min}(r_t(\theta)\hat{A}_{GAE,t},clip(r_t(\theta),1-\sigma,1+\sigma)\\
% &\hat{A}_{GAE,t})]
% \end{aligned}
% \end{equation}
% In this paper, we will primarily work with the $L^{CLIP+VF}$ surrogate loss in the algorithm implementation and TRPO in the analysis.
\subsection{Meta-Learned Hierarchies}
\begin{wrapfigure}{r}{0.4\textwidth}
\centering
\includegraphics[width=0.4\textwidth]{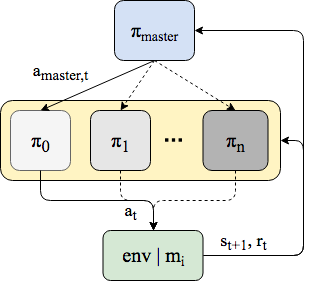}
\caption{A meta learning hierarchy similar to MLSH in ~\cite{frans2017meta}. The master is tasked with choosing a sub policy to maximize return in the current MDP $m_i$.}\vspace{-20pt}
\end{wrapfigure}
As a basis for our proposed MLAH framework, we consider a task with multiple objectives or latent states. In this context, we define a finite set of MDPs $\mathcal{M}$:$\{m_0,m,_1,\cdots,m_n\}$, where an MDP $m_i,\ i\in\{0, 1, \cdots, n\}$ is sampled for learning at time $t$. There exists a set of corresponding sub-policies $\Pi:\{\pi_0,\pi_1,\cdots,\pi_n\}$ which may individually be used at any instant. We then have $\mathcal{M} \rightarrow \mathcal{R}$ and define a joint hierarchal objective for $\mathcal{M}$ composed of sub-policies:
\begin{align}
\mathcal{R}(\Pi)=\mathbb{E}_{s_0,\pi_0,m_0...}\bigg[\sum_{t=0}^{T}\gamma^tr(s_t)|\, m_{i},\pi_{i} \bigg]
\end{align}
Every $m_i$ can be thought of as a unique objective in the same state-action space. In our case, the RL agent is not aware of the specific $m_i$ at time $t$. This could alternatively be thought of as a partially observable MDP (POMDP), however in this work we introduce a hierarchal RL architecture to explain the latent state.
%For completeness, the following definitions associated with the state-action value function %$Q_{\pi}$, the value function $V_{\pi}$, and the advantage function $A_{\pi}$ are given:
%\[
%\begin{split}
%   & Q_{\pi}(s_t,a_t)=\mathbb{E}_{s_{t+1},a_{t+1},...}\bigg[\sum_{l=0}^{\infty}\gamma^lr(s_{t+l})\bigg]\\
%\end{split}
%\]
%where $a_t\sim\pi(a_t|s_t), s_{t+1}\sim\mathcal{P}(s_{t+1}|s_t,a_t), \forall t\geq 0$.
%Thus, for the rest of paper, $\pi$ represents either the nominal or adversarial stochastic policy.
% First, we present an identity lemma based on \cite{schulman2015trust} and \cite{kakade2002approximately}, which indicates the expected discounted return of one policy $\hat{\pi}$ using the advantage over $\pi$.
This hierarchal framework depicted in Figure 1 has been presented in ~\cite{frans2017meta} as \textit{Meta-Learned Shared Hierarchies} (MLSH). $\pi_{master}$ describes an agent who's task is to select the appropriate sub-policy to maximize return. The master policy, $\pi_{master}$ receives the observed reward and environment state. This mapping is far easier to learn as apposed to re-learning each sub policy which may be re-used. Since each $m_i \in \mathcal{M}$ has a different $\mathcal{S} \rightarrow \mathcal{R}$ mapping, this makes $\pi_{master}$ have a non-stationary mapping across $\mathcal{S}$ which requires the parameters of $\pi_{master}$ to be reset on a predetermined interval.
%It is hypothesized that the advantage is a computationally simple and contextual metric which provides $\pi_{master}$ the necessary information to form a belief over the best sub-policy.
% For better quantifying the latter analysis, we formally give the definition of adversarial attack mathematically as follows.
% \begin{defn}
% There exists a constant $0\textless \sigma \leq 1$ such that a bounded adversarial attack $\mathbf{v}$ is an element of a set $\mathcal{D}$ which satisfies
% \begin{equation}
% \mathcal{D} = \{\mathbf{v}|\mathbf{v}\in\mathbb{R}^n; v_i\in[v_{min},v_{max}], i=1,2,...,n; Pr\{a_{worst}=\pi(a|s,\mathbf{v})\}\geq \sigma\}
% \end{equation}
% \end{defn}
\subsection{Adversary Models}
We consider adversaries that perturb the state provided to the agent at any given time instant. Formally,
\begin{defn}\label{defn1}
An adversarial attack is any possible state observation perturbation that leads the agent into incurring a sub-optimal return, which is less than the return of the learned optimal policy. In other words, $\mathcal{R}(\pi | attack)< \mathcal{R}(\pi)$. The adversary \textbf{may only} perturb the state observation channel, and not the reward channel itself.
\end{defn}
Note, when discussing adversarial attacks, a common practice is to mathematically define a feasible perturbation with respect to the observation space. This work presents an alternative approach (later in the analysis Section~\ref{analysis}) by focusing on expected frequency of attacks only and how it realizes in the RL decision space. This results in a framework which is more agnostic to a specific attack-model and  considers more than just the observation (data) space. However, it is important to note that the RL agent is not aware of any attack-model specifications.

%The following are simply used for technical discussions and analysis.
\section{Proposed Algorithms}
We begin with a brief motivation to the proposed Meta-Learned Advantage Hierarchy (MLAH) algorithm. An intelligent agent, such as a human with a set of skills, when presented with a new task, should try out one of the known skills or policies and examine its effectiveness. When the task changes, based on the expectation of usefulness of that skill, the agent may keep using the same skill or try another skill that may seem most appropriate for that task. In this context, given that the agent has developed accurate expectations of its sub-policies (skills), if the underlying task were to change at anytime, the agent may notice that the result of its action has changed with respect to what was expected. In an RL framework, comparing the expected return of a state to the observed return of some action is typically known as the \textit{advantage}. Therefore, such an advantage estimate can serve as a good indicator of underlying changes in a task that can be leveraged to switch from one sub-policy to another more appropriate sub-policy.
%This process of learning skills and their expected reward, while using those predictions to choose the next skill resembles that of an EM style optimization.
%More concretely, it is hypothesized that this meta-cognitive process we appear to experience with new tasks and partially observable environments may be emulated by regressing an advantage coordinate space which is formed by comparing the returns of states to the predicted values generated by each available sub-policy. The advantage may be seen as a highly contextual metric for evaluating the usefulness of each sub-policy and is stationary with respect to the latent states. 

With this motivation, we can map the current problem of learning policy under intermittent adversarial perturbations as a meta-learning problem. As our adversarial attacks (by definition ~\ref{defn1}) create a different state-reward map, a master policy may be able to detect an attack and help choose an appropriate sub-policy that corresponds to the adversarial scenario. More formally, we begin with two random policies that are meant to represent the two distinct partially observable conditions in our MDP, nominal states and adversarial perturbed states. One may begin by pre-training $\pi_{nom}$ in isolation seeing only nominal experiences. Since we can not assume or simulate the adversary, typically it is not possible to pre-train $\pi_{adv}$ and it must be left to $\pi_{master}$ to identify this alternative mapping. For each episode, we begin by collecting a trajectory of length T, allowing $\pi_{master}$ at every time step (or on an interval) to select a sub-policy to act based on the advantage coordinate observed. The advantage for $\pi_{master}$, represented by $\mathbf{A}_t$, can be calculated using only the previous state-reward or it can be computed as a generalized estimate over the past $h$ time-steps as a rolling window.
\begin{align}\label{eq_adv_coord}
\mathbf{A}_t =  \big[A_{GAE,t-h}|\pi_{nom},A_{GAE,t-h}|\pi_{adv}\big] \in \mathbb{R}^{2}\\
a_{master,t}=\pi_{*,t} = \underset{a}{\text{argmax}}\,\mathbb{E}_{s_t,\pi_i,m_i...}\bigg[\sum_{t=0}^{T}\gamma^tr(s_t,a)|\, m_{i}\bigg] \in \{ \pi_{nom},\pi_{adv} \}
\end{align}
\vspace{-15pt}

\begin{figure}[htbp]
\centering
\subfloat[Adversary interaction model]{\includegraphics[height=1.65in]{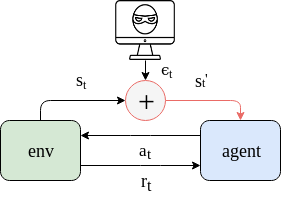}}\quad \quad
\subfloat[MLAH framework]{\includegraphics[height=1.95in]{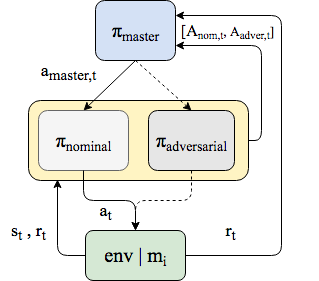}}
\caption{\textbf{a)} Illustration of the adversarial attack mechanism: corrupting the state observation, by injecting a perturbation $\epsilon$ before it reaches the agent, no perturbation in the reward signal. \textbf{b)} MLAH architecture: while similar to MLSH, key differences are: 1) master policy only observes the advantage of the sub-policy as a state and 2) only two sub-polices (nominal/adversarial) considered.}\vspace{-10pt}
\end{figure}

Observing the advantage over states and actions can be justified philosophically and has technical benefits when compared to other temporal state abstraction techniques that may be used to estimate the latent condition (RNN, LSTM). Although this mapping has potential to be noisy as the advantage can be trajectory dependent, it is static across the multiple sub-policies as opposed to a state-policy selection mapping which must be re-learned with every change in the latent condition.
\begin{algorithm}
\caption{MLAH}\label{mlah}
\SetKwInOut{Input}{Input}
%\begin{algorithmic}[1]
\Input{$\pi_{nom}$ and $\pi_{adv}$ sub-policies parameterized by $\theta_{nom}$ and $\theta_{adv}$; Master policy $\pi_{master}$ with parameter vector $\phi$.}
\text {Initialize $\theta_{nom},\,\theta_{adv},\,\phi$}\\
\For {pre-training iterations [optional]}
	{
	\text {Train $\pi_{nom}$ and $\theta_{nom}$ on only nominal experiences.}\\
	}
\For {learning life-time}
  {
  	\For {Time steps $t$ to $t + T$}
    {
    \text{Compute $\textbf{A}_t$ over sub-policies (see eq. \ref{eq_adv_coord})}\\
  	\text{select sub-policy to take action with $\pi_{master}$ using $\textbf{A}_t$ as observations}
    }
%    	\text{based on [$\hat{A}_{GAE,nom},\hat{A}_{GAE,adv}$] of the last N steps}\\
    \text {Estimate all $A_{GAE}$ for $\pi_{nom},\,\pi_{adv}$ over $T$}\\
   	\text {Estimate all $A_{GAE}$ for $\pi_{master}$ over $T$ with respect to $\textbf{A}_t$ observations}\\
	\text{Optimize $\theta_{nom}$ based on experiences collected from $\pi_{nom}$}\\
    \text{Optimize $\theta_{adv}$ based on experiences collected from $\pi_{adv}$}\\
    \text{Optimize $\phi$ based on all experiences with respect to $\textbf{A}_t$ observations}
   }
%\end{algorithmic}
\end{algorithm}

% The generalized advantage can be seen as a temporal abstraction of state that is highly efficient at representing the relative appropriateness of each sub-policy. It is a highly contextual metric in RL which is inexpensive alternative to complicated RNN RL agents which must learn full temporal state mappings.

\textbf{Advantage map as an effective metric to detect adversary:} To fool an RL agent into taking an alternative action, an adversary may use the policy network to compute a perturbation~\cite{goodfellow6572explaining}. For attack mitigation, the RNN-based visual-foresight method~\cite{lin2017detecting} is practical, considering the predicted policy distance from the chosen policy. However, it was reported~\cite{lin2017detecting} that such a scheme can be fooled with a series of likely state perturbations. However in MLAH, even if the adversary could compute a series of likely states to fool the agent, the advantage would still be affected and the master agent may detect the attack. The adversary would have to consecutively fool the agent with a state that would be expected to give an equally bad reward. This constraint would make the perturbation especially hard or infeasible to compute. We do acknowledge however that this method is slightly delayed such that the agent has to experience an off-trajectory reward before it can detect the adversary presence and may also have to observe long attack periods before learning the advantage mapping.
%%%%%%%%%%%%%%%%%%%%%%%%%%%%%%%%%%%%%%%%%%%%%%%%%%%%%%%%%%%%%%%%%%%%%%%%%%%%%%%%

\section{Analysis of Bias Mitigation and Policy Improvement}\label{analysis}
Here we present analysis to show that the proposed MLAH framework reduces bias in the value function baseline under adversarial attacks. We then show how reducing bias is inherently beneficial for policy learning (improvement in expected reward lower bound compared to the state-of-the-art as presented in ~\cite{schulman2015trust}) in the presence of adversaries. In order to estimate the expected value learned by a policy, we consider a first-order stochastic transition model (from nominal-$0$ to adversary-$1$ and vice versa) for the temporal profile of the attack as follows:
%, as shown in Figure~\ref{fig4}.
\begin{align*}
P=\begin{bmatrix}
p_{0|0} & p_{1|0}\\
p_{0|1} &p _{1|1}
\end{bmatrix} =
\begin{bmatrix}
m & 1-m \\
n & 1-n
\end{bmatrix}
\end{align*}
This defines a Markov chain ($p_{b|a}$ denotes the probability transitioning from $a$ to $b$). Let the stationary distribution for this Markov chain be denoted by, $v = [p_0, p_1]$ that satisfies $v = vP$. 
%We then solve the distribution analytically in terms of $m$ and $n$ in the following equations.
% \begin{equation}
% v = [p_0, p_1];\;v = vP,
% \end{equation}
%Solving this equation for $v$, we obtain
Therefore,
\begin{equation}
p_0 = \frac{n}{1-m+n}, \qquad p_1 = \frac{1-m}{1-m+n}
\end{equation}
which describes the long term expectation of visiting a nominal or adversarial state. As discussed in the preliminaries, trajectory experiences are handled with a distinct policy and value network when the adversarial attack is present. As the condition is perceived by the master agent, we can define two independent MDPs separately, i.e., one given a nominal state ($p_{\sim|0}$) and another given the perturbed state due to the adversary ($p_{\sim|1}$). 
%Based on $P$, it is immediately obtained that $p_{0|0} = m, p_{1|0} = 1-m;\;p_{0|1} = n, p_{1|1} = 1-n$. 
With this setup, we present an assumption as follows:
\begin{assumption}\label{assum1}
Long term expectation of visiting a nominal state is higher than that of adversarial state, i.e., for the stochastic transition model $P$, $n\textless m$.
\end{assumption}
% Because the condition is perceived by the meta agent, we can define two independent MDPs and new stationary distributions, one given a nominal state and another given the state has been perturbed by the adversary.
% \begin{align}
% p_{0|0} = m, \qquad p_{1|0}= 1-m\\
% p_{0|1} = n, \qquad p_{1|1}= 1-n
% \end{align}
% Suppose that an MDP has two latent conditions, i.e., 0 and 1. 
% \begin{defn}\label{defn2}
% $\mathbb{E}_{s \sim \mathcal{S}|0}V(s)$ is the expected discounted return over states $\mathcal{S}$ given that the policy \textbf{only} sees nominal conditions ($m=1$). Similarly, $\mathbb{E}_{s \sim \mathcal{S}|1}V(s)$ is the expected discounted return for the return of seeing \textbf{only} the adversarial conditioned states ($m=0,n=0$). For simplicity, $\mathbb{E}_{s \sim \mathcal{S}|0}V(s)$ is denoted by $V_0$ and $\mathbb{E}_{s \sim \mathcal{S}|1}V(s)$ is denoted by $V_1$.
% \end{defn}

Let $\mathbb{E}_{s \sim \mathcal{S}|0}V(s)$ be the expected discounted reward over states $\mathcal{S}$ given that the policy \textbf{only} sees nominal conditions ($m=1$). Similarly, let $\mathbb{E}_{s \sim \mathcal{S}|1}V(s)$ be the expected discounted reward for the policy when it sees the adversarial states ($m=0,n=0$) alone. We simplify the notations as follows: $\mathbb{E}_{s \sim \mathcal{S}|0}V(s) = V_0$ and $\mathbb{E}_{s \sim \mathcal{S}|1}V(s) = V_1$ as two \textit{value primitives}.

According to definition of the adversary (Definition~\ref{defn1}), we have $V_1 < V_0$ as a successful adversarial attack leads to a sub-optimal return. 
% Assume that we know the expected discounted returns for two conditioned MDPs that share the same state space but have a different latent condition. 
% Since there exists a stationary distribution, we can calculate the expected discounted return given the row-stochastic model for the condition ($m$ and $n$). 
We can now compare the expected discounted return for the \textit{unconditioned} and \textit{conditioned} learning scheme. 
%Based on the defined stochastic transition model $P$, 
Here, the unconditioned scheme refers to the learning scheme of a classical DRL agent with one policy. In this case, the expected discounted reward  under adversarial attacks can be expressed as:
\begin{align}\label{eq56}
\mathbb{E}_{unc,s \sim \mathcal{S}}V(s)=
V_0p_0+V_1p_1=V_0\frac{n}{1-m+n} + V_1\frac{1-m}{1-m+n}
\end{align}
On the other hand, the conditioned schemes refer to the two sub-policies (one given the nominal state and other given the adversarial state) based on the proposed MLAH framework. In this context, the expected discounted reward  conditioned on the nominal state under adversarial attacks can be expressed as:
\begin{align}\label{eq56}
\mathbb{E}_{\text{con},s \sim \mathcal{S}|0}V(s) = V_0p_{0|0} + V_1p_{1|0}=V_0 m + V_1(1-m)
\end{align}

%is referred to the stochastic model with transition probabilities of $p_0$ and $p_1$, and the 
%conditioned models are refereed to two stochastic models with transition probabilities, $p_{0|0},p_{0|1}$ and $p_{0|1},p_{1|1}$, respectively. It should be noted that the unconditioned model corresponds to the DRL agent with one policy and the conditioned model corresponds to our proposed MLAH framework.

%In this context, we discuss the difference between the unconditioned expected discounted return, $\mathbb{E}_{unc,s \sim \mathcal{S}}V(s)$, and the conditioned expected discounted return given a nominal state, $\mathbb{E}_{\text{con},s \sim \mathcal{S}|0}V(s)$. For the perturbed state by the adversary, one can follow the similar analysis procedure.

We now provide a lemma to compare the unconditioned and conditioned (given a nominal state) expected discounted rewards.
% We argue that the conditioned policy can outperform the unconditioned policy if and only if some condition is satisfied. One lemma is presented to introduce the condition that probabilities $m$ and $n$ should satisfy.
\begin{lem}\label{lem1}
Let Assumption~\ref{assum1} hold. $\mathbb{E}_{unc,s \sim \mathcal{S}}V(s)\textless \mathbb{E}_{\text{con},s \sim \mathcal{S}|0}V(s)$.
\end{lem}
See the proof in the Supplementary material. 

%\begin{rem}
% As it is expected that with the adversarial attack, the performance with conditioned policy should be better than that with the unconditioned policy, it is required that $n\textless m$ given that $V_1 \textless V_0$. 
% In a long term, the probability that a nominal state transits to another nominal state should be larger than the probability that an adversarial state transits to another adversarial state. Intuitively, such a condition practically makes sense as in the real-world problems, the frequency of occurrence of the adversarial attack is not supposed to be higher than the nominal condition. Let us justify the assumption that $V_1 \textless V_0$. Based on the definition of adversarial attack, it can be implied that during the adversarial states, the probability of “worst” actions resulting in the least $Q$ values is significantly higher than that in the nominal states. Therefore, in an expectation, we should have $V_1 < V_0$.
% \end{rem}
We next discuss different lower bounds of the expected discounted rewards for the conditioned and unconditioned policies. We begin with defining the observed bias in the \textit{state value} for both the conditioned and unconditioned policies by comparing the expected discounted reward to the original nominal value primitive $V_0$. Then, we have,
\[\delta_{con|0} = V_0 - \mathbb{E}_{\text{con},s \sim \mathcal{S}|0}V(s)=(1-m)(V_0-V_1),\quad
\delta_{unc} = V_0-\mathbb{E}_{unc,s \sim \mathcal{S}}V(s)=\frac{(1-m)(V_0-V_1)}{1-m+n}\] With this setup, we present the following lemma.
\begin{lem}\label{lem2}
Let Assumption~\ref{assum1} hold. $\delta_{con|0} < \delta_{unc}$.
\end{lem}
The proof is straightforward using Lemma~\ref{lem1} (see Supplementary material).

% While Lemma~\ref{lem2} shows that expected bias is reduced due to conditioning in our proposed framework, we note that the actual bias can be different from the expected bias. It stems from the fact that due to complex and uncertain environment, the true expected discounted return is not necessarily accessible.
%Due to the complex and uncertain environment, the true expected discounted return is not necessarily accessible and the actual expected discounted return is received such that there exits a bias between them. Although we have shown in Lemma~\ref{lem2} the relationship between the unconditioned and conditioned expected bias, for the convenience of analysis, we use $\delta$ to represent them in a general way. 
In this context, we express $V_0=\mathbb{E}_{con, s\sim \mathcal{S}|0}V(s)+\delta_{con|0}$ and $V_0=\mathbb{E}_{unc, s\sim \mathcal{S}}V(s)+\delta_{unc}$ in a general way as: $V(s)=\hat V(s)+\delta$, where $\delta$ is the observed bias in the state value. According to the definition of advantage function in Eq.~\ref{adv}, letting $\lambda=0$, we have $A_{\pi}(s_t,a_t)=r_t+\gamma V(s_{t+1})-V(s_t)$. Substituting $V(s)=\hat V(s)+\delta$ into the last equation yields
\begin{equation}
A_{\pi}(s_t,a_t)=r_t+\gamma\hat{V}(s_{t+1})-\hat{V}(s_t)+\gamma\delta_{s,t+1}-\delta_{s,t}=\hat{A}_{\pi}(s_t,a_t)+\gamma\delta_{s,t+1}-\delta_{s,t}
\end{equation}
where $\hat{A}_{\pi}(s_t,a_t)$ is the actual advantage function. While Lemma~\ref{lem2} shows that $\delta$ is reduced due to conditioning in our proposed framework, we note that the observed bias in the expected discounted reward can be different from that in the state value due to the complex and uncertain environment. 
%It stems from the fact that due to complex and uncertain environment, the true expected discounted reward is not necessarily accessible. 
Following the definition of the expected discounted reward in~\cite{schulman2015trust}, recalling $V(s)=\hat V(s)+\delta$, the relationship between true and actual expected discounted reward is: $\mathcal{R}(\pi)=\mathbb{E}_{s\sim\pi}[\hat{V}_{\pi}(s_t,a_t)+\delta]=\hat{\mathcal{R}}(\pi)+\hat{\delta}$, where $\hat{\delta}$ is observed bias in the expected discounted reward.
%, different from the expected bias $\delta$. 
We denote the observed bias in the reward for the unconditioned and conditioned cases as: $\hat{\delta}_{unc}$ and $\hat{\delta}_{con|0}$. Let $\Delta\hat{\delta} = \hat{\delta}_{unc} -\hat{\delta}_{con|0}$ and $\Delta\delta = \delta_{unc} -\delta_{con|0}$. We are now ready to discuss the lower bounds of the expected discounted rewards for the conditioned and unconditioned schemes. Before that, based on~\cite{schulman2015trust}, we introduce the maximum total variation divergence for any two different policies and use $\alpha$ to denote it for the rest of the analysis. We also first present one proposition to show the relationship between the actual expected discounted reward and its approximation. It is an extension of Theorem 1 in~\cite{schulman2015trust}, which helps characterize the main claim in the paper.
\begin{prop}\label{prop1}
Let Assumption~\ref{assum1} hold. Then the following inequality hold:
\begin{equation}\label{eq20}
\hat{\mathcal{R}}(\pi_{new})\geq\hat{L}_{\pi_{old}}(\pi_{new})-\frac{4\tilde{\epsilon}\gamma\alpha^2}{(1-\gamma)^2}
\end{equation}
where $\pi_{new}$ indicates the new policy, $\pi_{old}$ indicates the current policy, $\hat{L}_{\pi_{old}}(\pi_{new})=L_{\pi_{old}}(\pi_{new})+\delta-\hat{\delta}$, $L_{\pi_{old}}(\pi_{new})$ is the approximation of $\mathcal{R}(\pi_{new})$, i.e., $L_{\pi_{old}}(\pi_{new})=\mathcal{R}(\pi_{old})+\sum_{s}\rho_{\pi_{old}}(s)\sum_a\pi_{new}(a|s)A_{\pi_{old}}(s,a)$, $\rho$ is the discounted visitation frequencies as similarly defined in~\cite{schulman2015trust}, $\tilde{\epsilon}$ satisfies the following relationship
\begin{equation}
  \tilde{\epsilon}=\begin{cases}
    max_{s,a}|\hat{A}_{\pi}(s,a)| + (\gamma-1)\delta, & \text{if $\hat{A}_{\pi}(s,a)\geq (1-\gamma)\delta$}.\\
    -max_{s,a}|\hat{A}_{\pi}(s,a)| + (1-\gamma)\delta, & \text{if $0\textless\hat{A}_{\pi}(s,a)\textless (1-\gamma)\delta$}.\\
    max_{s,a}|\hat{A}_{\pi}(s,a)| + (1-\gamma)\delta, &\text{if $\hat{A}_{\pi}(s,a)\leq 0$}
  \end{cases}
\end{equation}
\end{prop}
See the proof in the supplementary material. We then arrive at the following result to show that using the conditioned policy allows to achieve a higher lower bound of expected discounted reward.
\begin{prop}\label{prop2}
If $\Delta\hat{\delta} < C\Delta V$, where $C \geq\frac{(m-n)(1-m)(4\gamma\alpha^2+1-\gamma)}{(1-m+n)(1-\gamma)}$ and $\Delta V=V_0 - V_1$, then the conditioned policy has a higher lower bound of expected discounted reward compared to that of the unconditioned policy.
% Let $\alpha=D^{max}_{TV}(\pi_{old},\pi_{new})$. If $\hat{A}_{\pi}(s,a)\geq (1-\gamma)\delta$ then the following inequalities hold:
% \begin{equation}
% \hat{L}_{\pi_{old}}(\pi_{new})_{unc}-\frac{4\hat{\epsilon}_{unc}\gamma\alpha^2}{(1-\gamma)^2}\textless \hat{L}_{\pi_{old}}(\pi_{new})_{con|0}-\frac{4\hat{\epsilon}_{con|0}\gamma\alpha^2}{(1-\gamma)^2}
% \end{equation}
\end{prop}
Detail development of the proposition along with the proof is presented in the Supplementary material.
\begin{rem} 
Proposition~\ref{prop1} suggests that under a certain condition, using the conditioned policy can improve the lower bound of the expected discounted return over the unconditioned policy. Intuitively, the condition demands the adversary to be sufficiently intelligent in order to have a large enough value for $\Delta V$. 
%When $\Delta\hat{\delta}\textgreater 0$ increases, the lower bound of $C\Delta V$ also becomes higher such that $\Delta V$ increases. Intuitively, if the adversary is intelligent enough so as to reduce $V$ sufficiently, the conditioned policy has a higher lower bound of expected discounted return as the unconditioned policy results in the higher $\hat{\delta}_{unc}$. When $\Delta\hat{\delta}\textless 0$, the condition can be easily satisfied as $\Delta V\textgreater 0$ and $C\textgreater 0$. However, this scenario leads to a contradiction to the conclusion of Lemma~\ref{lem1} so the above proposition implies that the proposed conditioned policy allows the agent to learn less biased.
\end{rem}
% The above Corollary states that if the advantage can satisfy some condition, the lower bound of expected discounted return for the conditioned policy is better than that of the unconditioned policy.
% \begin{rem}
% The condition above may seem constrictive, but it is possible to show that $\underset{s}{\text{max}}\,\hat{A}_{\pi}(s,a)$ is always $\geq \delta (1-\gamma)$. If we consider that in order to achieve a positive advantage, the value function must be biased to underestimate the return in the first place. So we must require that value function bias itself needs to be biased by \textit{atleast} $\delta (1-\gamma)$ in the first place. We can clearly say for any $\delta > 0$:
% \begin{align*}
% \delta \geq \delta(1-\gamma)\\
% 1 \geq (1-\gamma)
% \end{align*}
% Which is always true since $0 < \gamma < 1$. One can arrive at the same result for $\delta < 0$ when $\underset{s}{\text{max}}\,\hat{A}_{\pi}(s,a) < \delta (1-\gamma)$.
% \end{rem}
\section{Experimental Results}
In order to justify the theoretical implications of bias reduction using a conditioned policy optimization, we implemented the proposed framework introduced in Section 3 with a selection of simple adversary models. Because the meta-learned framework has many moving parts and can be subject to instabilities, we first consider a case where the master agent is an oracle in determining the presence of an adversary. Then we consider the advantage-based adversary detection by the master agent.
\subsection{Experimental Setup}
For all experiments, we use the proximal clipped objective $L(\theta)^{CLIP+VF}$ from ~\cite{schulman2017proximal} instead of a constrained trust region optimization in accordance with recent results showing similar performance and ease of implementation. We use the same optimization for the master agent, although we acknowledge this may not be the best method for only two action choices (nominal or adversarial), we propose this to generalize to an arbitrary number of sub-polices.
In every example, \textit{training} denotes the agent acting with an $\epsilon$-greedy exploration policy with adversarial attacks. Simultaneously, we run an \textit{evaluation} which executes a deterministic actions with the same policy, \textit{without adversarial attacks}, hence obtain much higher return values.
For the examples shown, we introduce the adversary on a fixed interval (e.g., $5000$ with adversary, $10000$ without). During that period, the adversary perturbs the state at \textit{every} time step. For page limit constraints, PPO parameters used in experiments such as deep network size and actor-batches can be found in the supplementary material.

\subsubsection{Stochastic $l_{\infty}$-bounded Attacks}
In this paper, for the purpose of experiment, we consider an attacker model that has the ability to perturb state information from the environment before it reaches the agent. Since gradient-based attacks for continuous action policies have not been thoroughly studied, the adversarial agent will not optimize it's attack for the agent's policy, but only sample the perturbation size and direction from a defined uniform distribution $\mathcal{U}(a,b)$ about the current state $\textbf{s}=[s_0,s_1,\cdots,s_n]$. This results in an attack where $s_{i,\, adversary} = s_{i} + \mathcal{U}(a,b)$ where the perturbation is bounded by the $l_{\infty}$ norm so that $\forall s_i \in \textbf{s} \quad \underset{i}{\text{max}}|s_{i} - s_{i,\,adversary}| \leq \epsilon_{attack}$. We find that this naive attack is effective enough to decrease the return of a policy. We specifically utilize white-noise attacks where $a=-b$ as well as bias attacks, where $a\neq b$ and $a<b$.

\subsection{Adversarial Bias Reduction with MLAH}
We begin by examining an RL environment where the master agent is asked to select the policy that corresponds to the current condition, i.e., nominal or adversarial. We acknowledge that this "policy" may not be the optimal master policy since a game may not be perfectly Markov. However, we find that this is sufficient to examine the policy improvement in some Openai Mujoco control environments~\cite{brockman2016openai}.
\begin{figure}[h]
\centering
\includegraphics[scale=0.43]{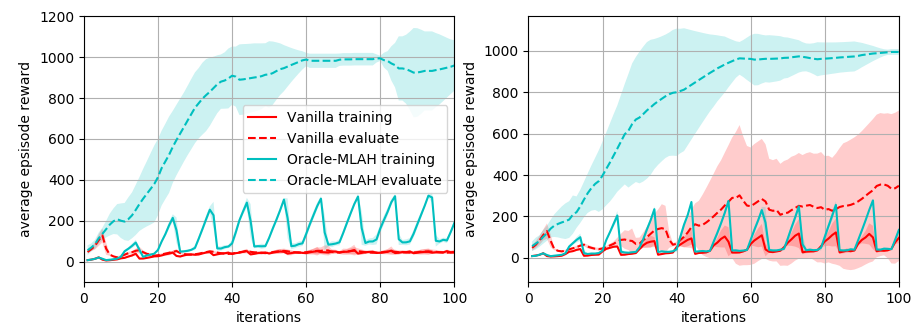}
\caption{Results of Oracle-MLAH and Vanilla PPO applied to the InvertedPendulum-v2 game with repeatedly scheduled attacks for $5000$ time steps and then off for $10000$, displaying a $1\sigma$ bound. \textbf{Left:} Case study with an extreme bias attack spanning the entire state-space. Vanilla policy is unable to resolve the correct mapping due to large disturbances in the state information, while MLAH improves nearly monotonically. \textbf{Right:} Case study with a weaker bias attack, Vanilla agent still struggles.}\label{ip_bias}\vspace{-10pt}
\end{figure}

\setlength\tabcolsep{2pt}
\begin{table}
\begin{center}
\begin{threeparttable}
\caption{Performance evaluation of Oracle-MLAH}\label{table1}
  \begin{tabular}{c|c|c|c|c}
    \hline
    &\multicolumn{2}{|c|}{\textbf{Normalized avg. training return}}&\multicolumn{2}{c}{\textbf{Normalized avg. evaluation return}} \\ \hline
    $m/n$& Vanilla& Oracle-MLAH& Vanilla& Oracle-MLAH\\ \hline
    $1.0/-$ & $0.96 \pm 0.03$& $0.96 \pm 0.03$ & 1.0 & $1.0$\\ \hline
    $0.995/0.005$ & $0.238 \pm .082$& $0.553 \pm 0.242$ & $0.471 \pm 0.051$ & $0.99 \pm 0.001$\\ \hline
    $0.95/0.05$ & $0.612 \pm .08$& $0.677 \pm 0.149$ & $0.644 \pm 0.078$ & $0.99 \pm 0.001$\\ \hline
    $0.8/0.2$ & $0.613 \pm 0.043$ & $0.728 \pm 0.063$ & $0.539 \pm 0.023$ & $0.994 \pm 0.165$\\ \hline
    $0.5/0.5$ & $0.749 \pm 0.093$ & $0.764 \pm 0.078$ & $0.787 \pm 0.010 $ &$0.948 \pm 0.086$\\
    \hline
  \end{tabular}
\begin{tablenotes}
    \small
    \item Comparison of the returns of Vanilla PPO and Oracle-MLAH under attacks over 40 policy optimization iterations with $1\sigma$ uncertainty bounds. The training return uses a stochastic policy for exploration and evaluation acts deterministically. The evaluation bias for the Oracle-MLAH remains substantially lower over all attack severity levels. Note when $m=n$, training returns are very similar as predicted by Eq. \ref{eq56}.
  \end{tablenotes}\vspace{-15pt}
\end{threeparttable}
\end{center}
\end{table}
The returns shown in Table \ref{table1} and Figure \ref{ip_bias} for long and intermittent bias attacks (large m and small n) clearly demonstrate the benefit of using distinct policies for nominal and adversarial states respectively. According to eq. \ref{eq56}, this attack condition produces the largest difference in bias between conditioned and unconditioned policies. As a policy can only solve for one state-action mapping and there are clearly two separate MDP state-reward distributions existing across time, a singly policy has no choice, but to optimize over the mean of these two distributions. Often times this results in not developing a useful policy for either condition as shown in figure \ref{ip_bias}. Enabling the use of multiple polices in this intermittent attack case allows the agent to optimize for both mappings, even learning to mitigate the reduced return during the adversarial attack. More simulation results using Open Gym environments such as \textit{MountainCarContinuous-v0} and \textit{Hopper-v2} ~\cite{brockman2016openai} are included in the supplementary material.

It can be seen in table \ref{table1} that as the switching expectations between nominal and adversarial states rise, the unconditioned (Vanilla) policy actually performs increasingly well, but still less than that of the conditioned (MLAH) policy. This is perhaps because the switching is quick enough to map the scenario to one state-reward distribution, which is favorable for a single policy agent. 

As anticipated by the analysis, when $m = n$, the training performances of both policies approach a similar value, however the conditioned MLAH agent was able to maintain a nearly unbiased evaluation return. This may be an artifact of the environment or adversary, which is relatively simple and unintelligent. Over longer attack periods, it may be unrealistic to expect the return to behave according to the stationary distribution expectation because the average resolves on a longer time scale than policy optimization.

\vspace{-10pt}
\begin{figure}[h]
\centering
\includegraphics[scale=0.49]{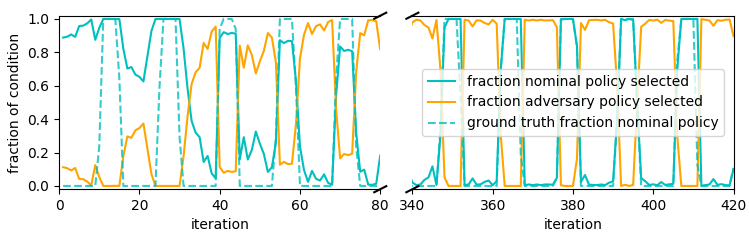}
\caption{Master agent's performance in learning from two random policies to decide which to employ to maximize the reward of \textit{InvertedPendulum-v2} with bounded $5000$ on, $10000$ off bias attacks. The master agent is not given any information on which states are perturbed by the adversary. After initial learning, the policy choices clearly diverge during the attack intervals with few exceptions.}\label{meta_switch}\vspace{-10pt}
\end{figure}

Next we put our master agent to the test, using the relative advantage coordinate mappings. This formulation is a novel alternative to previous meta-learned hierarchies which are non-stationary and need to be reset over time ~\cite{frans2017meta}. The relative advantage mapping is stationary across multiple MDPs under certain conditions. In order for the master agent to arrive at correct advantage-policy mapping, the policies themselves must also optimize to produce better advantage estimates in this expectation maximization (EM) type algorithm. This makes it challenging to produce a stable learning sequence of polices and advantage mappings. However, this mapping can be learned from ``nothing'' if an adversary creates a strong enough presence by altering the state-reward mapping (by Definition~\ref{defn1}). This optimization process is explained in more depth in the supplementary material. Depending on whether the nominal policy is pre-trained and the effectiveness of the adversary, the meta agent can reliably use each policy during the respective conditions. As seen in Figure~\ref{meta_switch}, an adversary is introduced in an intermittent manner and the master agent has two random sub-polices at its disposal. The agent optimizes to use one policy for the nominal and the other for the adversarial conditions to optimize its reward. The policy-selection results in Figure \ref{meta_switch} may resemble a Bayesian non-parametric latent state estimator~\cite{fox2011bayesian}. However, being entirely in the context of RL, MLAH is unique and uses the advantage observation and a meta-learning objective to form a belief over the latent conditions.   
% \begin{figure}[ht]
% \includegraphics[scale=0.7]{./figs/meta_return.png}
% \caption{Returns of successful meta-learned hierarchy from nothing.}
% \end{figure}
%%%%%%%%%%%%%%%%%%%%%%%%%%%%%%%%%%%%%%%%%%%%%%%%%%%%%%%%%%%%%%%%%%%%%%%%%%%%%%%%
\section{Conclusions}
We have discussed a new MLAH framework for handling adversarial attacks in an online manner specifically in the context of RL. This framework is attack-model agnostic and presents a general way of examining adversarial attacks in the temporal domain. Analyzing the hierarchical policy MLAH in this way, we can show that under certain conditions, the return lower-bound is improved when compared to a single policy agent. In future research, we aim to improve the stability of MLAH by optimizing the master agent function, perhaps using a more simple method to regress the advantage space. We will also attempt to extend MLAH to a more general framework for decision problems with multiple time-varying objectives.

%%%%%%%%%%%%%%%%%%%%%%%%%%%%%%%%%%%%%%%%%%%%%%%%%%%%%%%%%%%%%%%%%%%%%%%%%%%%%%%%

%%%%%%%%%%%%%%%%%%%%%%%%%%%%%%%%%%%%%%%%%%%%%%%%%%%%%%%%%%%%%%%%%%%%%%%%%%%%%%%%
\clearpage
\bibliographystyle{unsrt}
\bibliography{safe_rl}

%\begin{thebibliography}{99}

% \bibitem{trpo}
% Schulman, J., Levine, S., Moritz, P., Jordan, M.~I., \& Abbeel, "Trust Region Policy Optimization", P.\ 2015, arXiv:1502.05477
% \bibitem{ppo}
% Schulman, J., Wolski, F., Dhariwal, P., Radford, A., \& Klimov, "Proximal Policy Optimization Algorithms", O.\ 2017, arXiv:1707.06347
% \end{thebibliography}

\newpage

% \section{Supplementary Materials for ``Online Robust Policy Learning in the Presence of Unknown Adversaries"}\label{supp}
\section{Supplementary Materials}\label{supp}
\subsection{Additional Analysis}
This section presents the analysis for all of lemmas and propositions and additional analysis.

\textbf{Transition Mechanism of Adversary MDP}:

\begin{figure}[h]
\centering
\includegraphics[width=0.35\textwidth]{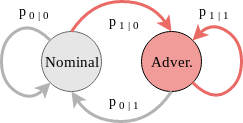}
\caption{Assumed mechanism (only for analysis purpose) for nominal to adversary state transitions: $p_{0|0}$ signifies the probability that a nominal state transits to another nominal state; $p_{1|0}$ signifies the probability that a nominal state transits to an adversarial state; $p_{0|1}$ signifies the probability that an adversarial state transits to a nominal state; $p_{1|1}$ signifies the probability that an adversarial state transits to another adversarial state}\label{fig4}
\end{figure}

The rest of the analysis in this section is based on the above transition mechanism.

\textbf{Proof of Lemma~\ref{lem1}}:
\begin{proof}
Based on the definitions of $\mathbb{E}_{unc,s \sim \mathcal{S}}V(s)$ and $\mathbb{E}_{con,s \sim \mathcal{S}|0}V(s)$, we have\[\mathbb{E}_{unc,s \sim \mathcal{S}}V(s)-\mathbb{E}_{con,s \sim \mathcal{S}|0}V(s)=V_0\frac{n}{1-m+n}+V_1\frac{1-m}{1-m+n}-V_0m-V_1(1-m)\]
With some mathematical manipulation, we have
\begin{equation}
\mathbb{E}_{unc,s \sim \mathcal{S}}V(s)-\mathbb{E}_{con,s \sim \mathcal{S}|0}V(s)=\frac{(V_0-V_1)(n-m)(1-m)}{1-m+n}
\end{equation}
As $V_1\textless V_0$ and $n\textless m$, then we get the desired results.
\end{proof}

\textbf{Proof of Lemma~\ref{lem2}}: 
\begin{proof}
As $V_1\textless V_0$, then $V_0-V_1\textgreater 0$. Based on the definitions of $\delta_{con|0}$ and $\delta_{unc}$, and Lemma~\ref{lem1}, the desired result is immediately obtained.
\end{proof}

The following analysis is for establishing the relationship between the true and actual expected discounted rewards.

For completeness, we rewrite or redefine some definitions here to characterize the analysis. We denote by $\hat{V}(s)$ the actual state value of the learned policy (i.e., the conditioned or unconditioned). Define the relationship between the true state value and actual state value as:
\[V(s) = \hat{V}(s) + \delta\]
which can be adaptive to the unconditioned or conditioned policy by substituting different bias. $\delta$ is the observed bias in the state value. We also denote by $\pi$ and $\tilde{\pi}$ the current policy and the new policy. According to the definition of advantage function in Eq.~\ref{adv}, letting $\lambda = 0$, we have
\[A_{\pi}(s_t, a_t) = r_t + \gamma V(s_{t+1}) - V(s_t)\]
Substituting $V(s) = \hat{V}(s) + \delta$ into the last equation yields
\begin{equation}\label{eq10}
A_{\pi}(s_t, a_t)=r_t + \gamma\hat{V}(s_{t+1}) - \hat{V}(s_t) + \gamma\delta_{s,t+1} - \delta_{s,t} = \hat{A}_{\pi}(s_t, a_t) + \delta_{s,t+1}(\gamma - 1)
\end{equation}
where $\hat{A}_{\pi}(s_t, a_t)$ is the actual advantage function under a learned policy. Based on the definition of expected discounted reward in~\cite{schulman2015trust}, we have 
\begin{equation}
\mathcal{R}(\pi) =\mathbb{E}_{s\sim\pi}\bigg[V_{\pi}(s_t,a_t)\bigg]
\end{equation}
which results in the relationship between the true and actual expected discounted rewards as follows
\begin{equation}\label{return}
\mathcal{R}(\pi) = \mathbb{E}_{s\sim\pi}\bigg[\hat{V}_{\pi}(s_t,a_t)+\delta\bigg] = \mathcal{\hat{R}}(\pi) +  \hat{\delta}
\end{equation}
% One may argue why the $\hat{\delta}$ is not equal to 0 as in~\cite{schulman2015trust}. However, in this context, we consider the adversarial attacks such that the states are not always nominal. 
where $\hat{\delta}$ is the observed bias in the expected discounted reward. It is immediately obtained that corresponding to different learned policies, $\hat{\delta}$ is not the same. In this context, we define $\hat{\delta}_{unc}$ as the observed bias in the expected discounted reward caused by the unconditioned policy and $\hat{\delta}_{con|0}$ as the observed bias in the expected discounted reward caused by the conditioned policy. Now we analyze the expected discounted reward of the new policy $\tilde{\pi}$ in terms over the current policy $\pi$ in order to know the difference between different policies during the learning process. Following~\cite{schulman2015trust}, we define the expected discounted reward of $\tilde{\pi}$ as follows
\begin{equation}\label{eq13}
\mathcal{R}(\tilde{\pi})=\mathcal{R}(\pi)+\mathbb{E}_{s,a\sim\hat{\pi}}\bigg[\sum_{t=0}^T\gamma^tA_{\pi}(s_t,a_t)\bigg]
\end{equation}
% Similarly,
% the expected return of another policy $\tilde{\pi}$ in terms of the advantage of $\pi$ is
% \begin{equation}\label{return_1}
% \mathcal{R}(\tilde{\pi}) = \mathcal{R}(\pi) + \mathbb{E}_{s,a\sim\tilde{\pi}}\bigg[\sum_{t=0}^T\gamma^tA_{\pi}(s_t,a_t)\bigg]
% \end{equation}
Hence, combining Eq.~\ref{eq10} and Eq.~\ref{eq13} we obtain the expected discounted reward of the new policy $\tilde{\pi}$ with respect to the expected discounted reward of the current policy $\pi$, the actual advantage and the observed bias in the state value.
\begin{equation}\label{eq14}
\mathcal{R}(\tilde{\pi}) =\mathcal{R}(\pi)+\mathbb{E}_{s,a\sim\tilde{\pi}}\bigg[\sum_{t=0}^T\gamma^t\hat{A}_{\pi}(s_t,a_t)\bigg]+ \mathbb{E}_{s,a\sim\tilde{\pi}}\bigg[\sum_{t=0}^T\gamma^t(\gamma\delta_{s,t+1} - \delta_{s,t})\bigg]
\end{equation}
As we use the same neural networks to estimate the actual state values, we assume that in Eq.~\ref{eq14} the expectation of bias $\delta_{s,t}$ given the state $s$ can be treated equally as constant, represented by $\delta$ for convenience of analysis. 
% in In this context, we assume that the expectation of bias is constant such that $\delta_{s,t+1}\approx\delta_{s,t}=\delta$ due to the same estimator network used. 
Therefore, by substituting Eq.~\ref{return} the last equality becomes as follows
\begin{equation}\label{true_act_ret}
\mathcal{R}(\tilde{\pi}) = \mathcal{\hat{R}}(\pi) + \hat{\delta} +\mathbb{E}_{s,a\sim\tilde{\pi}}\bigg[\sum_{t=0}^T\gamma^t\hat{A}_{\pi}(s_t,a_t)\bigg] - \delta = \mathcal{\hat{R}}(\tilde{\pi}) + \hat{\delta} - \delta
\end{equation}
which shows the true expected discounted reward of the policy $\tilde{\pi}$ with respect to its actual expected discounted reward $\mathcal{\hat{R}}(\tilde{\pi})$, the observed bias in the expected discounted reward, $\hat{\delta}$, and the observed bias in the state value, $\delta$.

For the rest of analysis, we follow the similar analysis procedure presented~\cite{schulman2015trust} and for convenience we denote by $\pi_{old}$ the current policy $\pi$ and by $\pi_{new}$ the new policy $\tilde{\pi}$. Following~\cite{schulman2015trust}, we first rewrite Eq.~\ref{eq13} as the following equation
\begin{equation}
\begin{split}
\mathcal{R}(\pi_{new}) &= \mathcal{R}(\pi_{old})+\sum_{s}\sum_{t=0}^T\gamma^t\mathcal{P}(s_t=s|\pi_{new})\sum_a\pi_{new}(a|s)A_{\pi_{old}}(s,a)\\
&=\mathcal{R}(\pi_{old})+\sum_{s}\rho_{\pi_{new}}(s)\sum_a\pi_{new}(a|s)A_{\pi_{old}}(s,a)
\end{split}
\end{equation}
where $\rho_{\pi_{new}}$ is the discounted visitation frequencies as similarly defined in~\cite{schulman2015trust}. Then we define an approximation of $\mathcal{R}(\pi_{new})$ as
\begin{equation}\label{eq17}
L_{\pi_{old}}(\pi_{new}) = \mathcal{R}(\pi_{old})+\sum_{s}\rho_{\pi_{old}}(s)\sum_a\pi_{new}(a|s)A_{\pi_{old}}(s,a)
\end{equation}
due to the complex dependence of $\rho_{\pi_{new}}$ on $\pi_{new}$. Similarly, according to Eq.~\ref{true_act_ret} we have
\begin{equation}\label{eq18}
L_{\pi_{old}}(\pi_{new}) = \hat{L}_{\pi_{old}}(\pi_{new}) + \hat{\delta} - \delta
\end{equation}
For completeness, we state the main theorem from~\cite{schulman2015trust} to guarantee the monotonic improvement. Before that, we need to define the total variation divergence for two different discrete probability distributions $q,o$, i.e.,
$D_{TV}(q||o)=\frac{1}{2}\sum_i|q_i-o_i|$,
based on which, we define
$D^{max}_{TV}(\pi_{old},\pi_{new})=max_sD_{TV}(\pi_{old}(\cdot|s)||\pi_{new}(\cdot|s))$.
Following~\cite{schulman2015trust},
we state the main theorem from~\cite{schulman2015trust} to guarantee the monotonic improvement.
\begin{thm}{(Theorem 1 in~\cite{schulman2015trust})}\label{thm1}
Let $\alpha=D^{max}_{TV}(\pi_{old},\pi_{new})$. Then the following bound holds:
\begin{equation}\label{eq19}
\mathcal{R}(\pi_{new})\geq L_{\pi_{old}}(\pi_{new})-\frac{4\epsilon\gamma\alpha^2}{(1-\gamma)^2}
\end{equation}
where $\epsilon=max_{s,a}|A_\pi(s,a)|$.
\end{thm}
With this, we arrive at the following proposition to demonstrate the relationship between the actual expected discounted reward and its approximation.

\textbf{Proposition~\ref{prop1}}
Let $\alpha=D^{max}_{TV}(\pi_{old},\pi_{new})$. Then the following inequality hold:
\begin{equation}\label{eq20}
\hat{\mathcal{R}}(\pi_{new})\geq\hat{L}_{\pi_{old}}(\pi_{new})-\frac{4\tilde{\epsilon}\gamma\alpha^2}{(1-\gamma)^2}
\end{equation}
where $\tilde{\epsilon}$ satisfies the following relationship
\begin{equation}
  \tilde{\epsilon}=\begin{cases}
    max_{s,a}|\hat{A}_{\pi}(s,a)| + (\gamma-1)\delta, & \text{if $\hat{A}_{\pi}(s,a)\geq (1-\gamma)\delta$}.\\
    -max_{s,a}|\hat{A}_{\pi}(s,a)| + (1-\gamma)\delta, & \text{if $0\textless\hat{A}_{\pi}(s,a)\textless (1-\gamma)\delta$}.\\
    max_{s,a}|\hat{A}_{\pi}(s,a)| + (1-\gamma)\delta, &\text{if $\hat{A}_{\pi}(s,a)\leq 0$}
  \end{cases}
\end{equation}
\begin{proof}
Combining Eq.~\ref{eq18} with the proof of Lemmas 1, 2, and 3 in~\cite{schulman2015trust}, we can arrive at the similar form of conclusion as shown in Theorem~\ref{thm1}. The difference between the conclusion in Theorem~\ref{thm1} and Proposition~\ref{prop1} is when we consider the actual expected discounted reward, the $\tilde{\epsilon}$ value is different from the $\epsilon$ value in Eq.~\ref{eq19}. We next discuss the new value for $\tilde{\epsilon}$. As the advantage function has the following relationship
\[A_{\pi}(s_t, a_t)=\hat{A}_{\pi}(s_t, a_t) + \delta(\gamma-1)\] Then, $\tilde{\epsilon} = max_{s,a}|\hat{A}_{\pi}(s,a) + (\gamma-1)\delta|$. Since $\delta\textgreater 0$ and $\gamma-1\textless 0$, we need to discuss the sign of $\hat{A}_{\pi}(s,a) + (\gamma-1)\delta$. Three cases are discussed as below:
\begin{enumerate}
\item When $\hat{A}_{\pi}(s,a) + (\gamma-1)\delta \geq 0$ such that $\hat{A}_{\pi}(s,a)\geq (1-\gamma)\delta$, $\tilde{\epsilon} = max_{s,a}|\hat{A}_{\pi}(s,a)| + (\gamma-1)\delta$,
\item When $\hat{A}_{\pi}(s,a) + (\gamma-1)\delta \leq 0$ and if $0\textless\hat{A}_{\pi}(s,a)\textless (1-\gamma)\delta$, $\tilde{\epsilon} = -max_{s,a}|\hat{A}_{\pi}(s,a)| + (1-\gamma)\delta$,
\item When $\hat{A}_{\pi}(s,a) + (\gamma-1)\delta \leq 0$ and if $\hat{A}_{\pi}(s,a)\leq 0$, $\tilde{\epsilon} = max_{s,a}|\hat{A}_{\pi}(s,a)| + (1-\gamma)\delta$,
\end{enumerate}
which completes the proof.
\end{proof}
\begin{rem}
The condition $\hat{A}_{\pi}(s,a)\geq (1-\gamma)\delta$ above may seem constrictive, but it can hold. If we consider that in order to achieve a positive advantage, the value function must be biased to underestimate the reward at the beginning. Therefore, the value function bias itself needs to be biased by \textit{at least} $\delta(1-\gamma)$ at the beginning. Hence, for any $\delta\textgreater 0$, we have $\delta(1-\gamma)\textless\delta$, which is always true as $0\textless\gamma\textless 1$. One can arrive at the same result for $\delta\textless 0$ when $\hat{A}_{\pi}(s,a)\leq (1-\gamma)\delta$.
\end{rem}
% According to Eq.~\ref{true_act_ret}, and we have defined the approximation of $\mathcal{R}(\pi_{new})$ as $L_{\pi_{old}}(\pi_{new})$, we can attain the following relationship similarly: 

Now we will show the Proposition~\ref{prop1} with the condition $\hat{A}_{\pi}(s,a)\geq (1-\gamma)\delta$. 

\textbf{Proof of Proposition~\ref{prop2}}: 
\begin{proof}
For assessing the new lower bound, we have exactly accounted for the bias in both conditioned and unconditioned policies. Therefore, according to Theorem~\ref{thm1}, Eq.~\ref{eq18}, and Eq.~\ref{eq20}, we have
\begin{equation}\label{eq22}
\begin{split}
L_{\pi_{old}}(\pi_{new})-\frac{4\epsilon\gamma\alpha^2}{(1-\gamma)^2}&=\bigg(\hat{L}_{\pi_{old}}(\pi_{new})\bigg)_{con|0}+\hat{\delta}_{con|0}-\delta_{con|0}-\frac{4\tilde{\epsilon}_{con|0}\gamma\alpha^2}{(1-\gamma)^2}\\
&=\bigg(\hat{L}_{\pi_{old}}(\pi_{new})\bigg)_{unc}+\hat{\delta}_{unc}-\delta_{unc}-\frac{4\tilde{\epsilon}_{unc}\gamma\alpha^2}{(1-\gamma)^2}
\end{split}
\end{equation}
The $\bigg(\hat{L}_{\pi_{old}}(\pi_{new})\bigg)_{con|0}$ and $\bigg(\hat{L}_{\pi_{old}}(\pi_{new})\bigg)_{unc}$ signify the approximation of $\mathcal{\hat{R}}(\pi_{new})$ in both conditioned and unconditioned policies, respectively. Similarly, $\tilde{\epsilon}_{con|0}$ and $\tilde{\epsilon}_{unc}$ indicate the different upper bounds corresponding to the conditioned and unconditioned policies, respectively. Let $\hat{\epsilon} = max_{s,a}|\hat{A}_{\pi}(s,a)|$ such that we have $\hat{\epsilon}_{con|0}$ and $\hat{\epsilon}_{unc}$ for the conditioned and unconditioned policies. Due to the condition that $\hat{A}_{\pi}(s,a)\geq (1-\gamma)\delta$, based on Proposition~\ref{prop1} we have 
\begin{equation}\label{eq23}
\tilde{\epsilon}_{unc} = \bigg(max_{s,a}|\hat{A}_{\pi}(s,a)|\bigg)_{unc} + (\gamma-1)\delta_{unc} = \hat{\epsilon}_{unc} + (\gamma-1)\delta_{unc}
\end{equation}
and
\begin{equation}\label{eq24}
\tilde{\epsilon}_{con|0} = \bigg(max_{s,a}|\hat{A}_{\pi}(s,a)|\bigg)_{con|0} + (\gamma-1)\delta_{con|0} = \hat{\epsilon}_{con|0} + (\gamma-1)\delta_{con|0}
\end{equation}
Hence, substituting Eq.~\ref{eq23} and Eq.~\ref{eq24} into Eq.~\ref{eq22}, we have
\begin{equation}\label{eq25}
\begin{split}
&\bigg(\hat{L}_{\pi_{old}}(\pi_{new})\bigg)_{con|0}+\hat{\delta}_{con|0}-\delta_{con|0}-\frac{4\hat{\epsilon}_{con|0}\gamma\alpha^2}{(1-\gamma)^2}+\frac{4\delta_{con|0}\gamma\alpha^2}{1-\gamma}\\
&=\bigg(\hat{L}_{\pi_{old}}(\pi_{new})\bigg)_{unc}+\hat{\delta}_{unc}-\delta_{unc}-\frac{4\hat{\epsilon}_{unc}\gamma\alpha^2}{(1-\gamma)^2}+\frac{4\delta_{unc}\gamma\alpha^2}{1-\gamma}
\end{split}
\end{equation}
By the condition that $\Delta\hat{\delta}\textless C\Delta V$ and $\Delta\hat{\delta}=\hat{\delta}_{unc}-\hat{\delta}_{con|0}$, we have
\begin{equation}
\begin{split}
\hat{\delta}_{unc}-\hat{\delta}_{con|0}&\textless \frac{(m-n)(1-m)(4\gamma\alpha^2+1-\gamma)}{(1-m+n)(1-\gamma)}\Delta V\\
&=\Delta V\frac{(m-n)(1-m)}{1-m+n}\frac{4\gamma\alpha^2+1-\gamma}{1-\gamma}\\
&=\Delta V\frac{m-n+mn-m^2+1-m+m-1}{1-m+n}\frac{4\gamma\alpha^2+1-\gamma}{1-\gamma}\\
&=\Delta V\frac{1-m-(1-m+n-m+m^2-mn)}{1-m+n}\frac{4\gamma\alpha^2+1-\gamma}{1-\gamma}\\
&=\Delta V\frac{1-m-(1-m)(1-m+n)}{1-m+n}\frac{4\gamma\alpha^2+1-\gamma}{1-\gamma}\\
&=\bigg(\frac{(1-m)\Delta V}{1-m+n}-(1-m)\Delta V\bigg)\bigg(\frac{4\gamma\alpha^2}{1-\gamma}+1\bigg)\\
\end{split}
\end{equation}
According to the definition of bias for the expected discounted reward, we have
\begin{equation}
\begin{split}
\hat{\delta}_{unc}-\hat{\delta}_{con|0}&\textless(\delta_{unc}-\delta_{con|0})\bigg(\frac{4\gamma\alpha^2}{1-\gamma}+1\bigg)\\
&=\frac{4\gamma\delta_{unc}\alpha^2}{1-\gamma}-\frac{4\gamma\delta_{con|0}\alpha^2}{1-\gamma}+\delta_{unc}-\delta_{con|0}
\end{split}
\end{equation}
The last inequality yields the following relationship:
\begin{equation}
\frac{4\gamma\delta_{con|0}\alpha^2}{1-\gamma}+\delta_{con|0}-\hat{\delta}_{con|0}\textless \frac{4\gamma\delta_{unc}\alpha^2}{1-\gamma}+\delta_{unc}-\hat{\delta}_{unc}
\end{equation}
which results in the next inequality, combined with Eq.~\ref{eq25}
\begin{equation}
\bigg(\hat{L}_{\pi_{old}}(\pi_{new})\bigg)_{con|0}-\frac{4\hat{\epsilon}_{con|0}\gamma\alpha^2}{(1-\gamma)^2}\textgreater \bigg(\hat{L}_{\pi_{old}}(\pi_{new})\bigg)_{unc}-\frac{4\hat{\epsilon}_{unc}\gamma\alpha^2}{(1-\gamma)^2}
\end{equation}
which suggests that by the conditioned policy, the lower bound of expected discounted reward is higher. It completes the proof.
\end{proof}
\subsection{Meta Optimization of the Advantage Space}
\begin{figure}[h]
\centering
\includegraphics[scale=0.27]{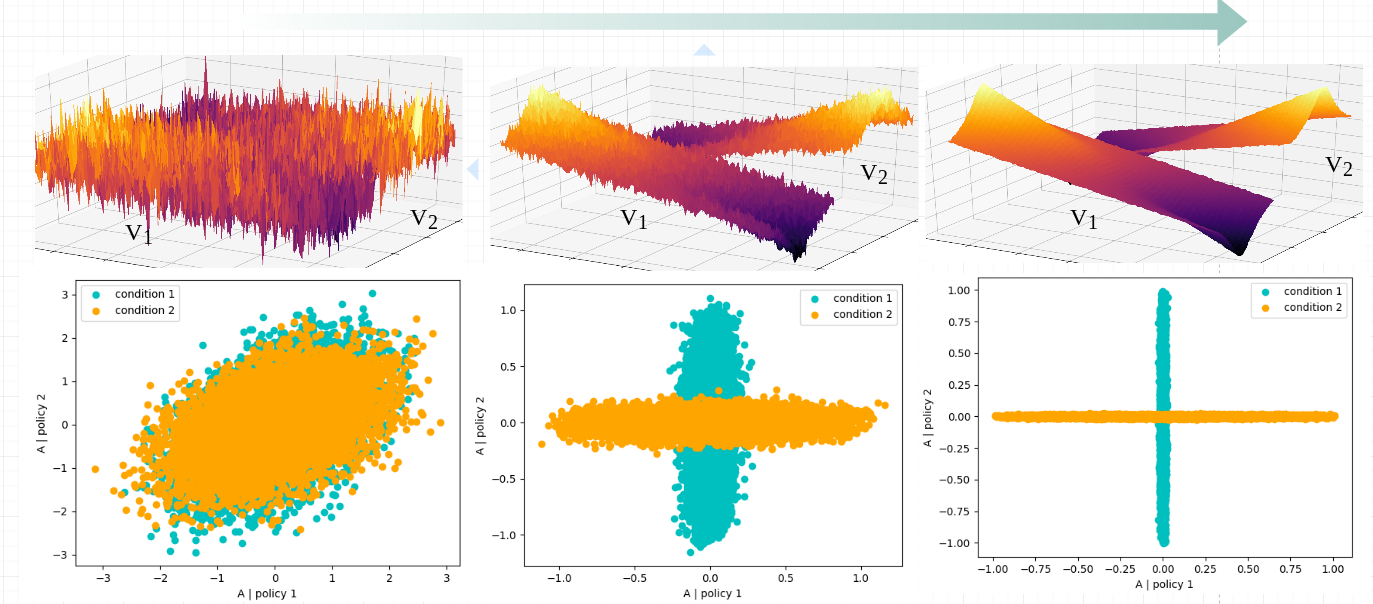}
\caption{Depicted is the visualized meta optimization process for a multi-objective game. In each frame there are two surface, each representing the belief of the state-value map. As the map becomes more accurate, the advantage space becomes easier to regress, which improves value accuracy and so on. One can draw parallels to some sort of expectation maximization style algorithm.}\label{evolve}
\end{figure}

To better explain the use of the advantage coordinate space, we provide some additional illustrations of the interesting optimization process at play. For visualization purposes, in figure \ref{evolve} we simulated a value surface with injected noise for a game in which there are two goal positions on a 2D plane, one at [$-1,0$] and the other at [$1,0$]. At any moment the goal may be at only one of these positions. When we create two polices to learn each distinct goal and value surface, we start from nothing, and the advantage space extremely noisy. The master agent will try its best to select sub-policies given this mapping and incrementally, each policy will become slightly better, meaning there is less bias and variance in the value function predictor. This will then allow the master agent to select policies with even greater accuracy, in result, improving the two value function accuracy more. One can see from this iterative process that it can hopefully achieve both an accurate master, and high-performing distinct polices simultaneously. This is an interesting way to perform an EM style optimization because it is \textit{only} defined by a reward signal. All other optimization steps can be derived from that single scalar signal.
\subsection{Additional Results}
\begin{figure}[h]
\centering
\includegraphics[scale=0.55]{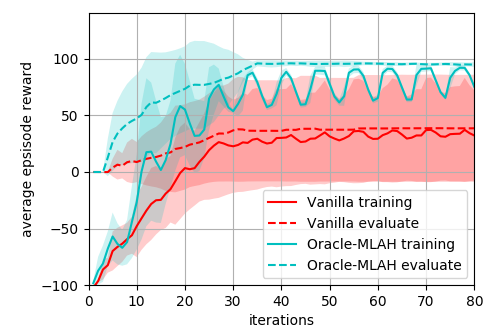}
\caption{Here we show an adversary that is implements strong stochastic white noise on the MountainCarContinuous-v0 environment. The baseline for the adversarial transitions happens to be approximately $5.0$ which makes our $\Delta V$ about $90.0$ reward points. According to the approximate $m$ and $n$ for this experiment, the bias should be about $60.0$ reward points ($30.0$ return), which is approximately the mean return for the Vanilla policy.}\label{mc}
\end{figure}

\begin{figure}[h]
\centering
\includegraphics[scale=0.5]{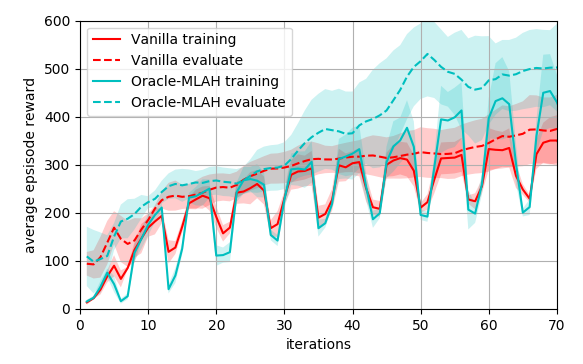}
\caption{Here we show an adversary that is implements strong stochastic white noise on the Hopper-v2 environment. This environment-adversary pair is particularly interesting because it shows that the unconditioned policy actually learned to handle the adversary more effectively than the conditioned MLAH. However, it obviously suffers in the nominal condition, while MLAH receives significantly higher returns.}\label{hopper}
\end{figure}

In this section, we provide some additional experiments, results and illustrations that may help the reader better understand the implications of the paper. We have tested the oracle-MLAH based protocol on several Gym environments and of these we show \textit{MountainCarContinuous-v0} Figure \ref{mc} and \textit{Hopper-v2} Figure \ref{hopper} case studies with white noise attacks and discuss some interesting observations made in each. All experiment are once again using the PPO clipped objective function with value prediction bonus.
\subsubsection{MountainCarContinuous-v0}
This experiment using the \textit{MountainCarContinuous-v0} environment was particularly insightful due to the behavior of the bias. The Vanilla policy was able to achieve considerable reward and the difference between it's nominal and adversarial peaks was small. It can be noted that out of the average maximum nominal reward of $\approx 95.0$ and a minimum with adversaries $\approx 5.0$, the expected biased return should have been $30.0$ according to $\Delta V$ from eq. \ref{eq56}. This is approximately the observed average return for the Vanilla policy.
\subsubsection{Hooper-v2}
The \textit{Hopper-v2} environment behaved similar to others during the attacks, except that the average performance during the attack appeared to be lower for the MLAH oracle, however MLAH oracle remained less biased in the nominal case. This is a curious observation that tells us that MLAH may not always mitigate the attack as well as a single Vanilla policy (in this case PPO).
% \begin{figure}{scale=0.5}
% \centering
% \includegraphics[width=0.35\textwidth]{./figs/adv_mdp.png}
% \caption{adversary MDP figure caption \color{red} remove the figure from main text\color{black}}\label{fig4}
% \end{figure}
\end{document}